\documentclass[conference]{IEEEtran}
\usepackage{cite}
\usepackage{amsmath,amssymb,amsfonts}
\usepackage{listings}
\usepackage{algorithmic}
\usepackage{graphicx}
\usepackage{textcomp}
\usepackage{xcolor}
\usepackage{hyperref}
\usepackage[normalem]{ulem}
\useunder{\uline}{\ul}{}
\usepackage [autostyle, english = american]{csquotes}
\MakeOuterQuote{"}

\graphicspath{ {images/} }
\def\BibTeX{{\rm B\kern-.05em{\sc i\kern-.025em b}\kern-.08em
    T\kern-.1667em\lower.7ex\hbox{E}\kern-.125emX}}

\begin{document}

\title{Recognition of Handwritten Japanese Characters Using Ensemble of Convolutional Neural Networks}
\author{\IEEEauthorblockN{Angel I. Solis\IEEEauthorrefmark{1},
Justin Zarkovacki\IEEEauthorrefmark{2}, John Ly\IEEEauthorrefmark{3} and
Adham Atyabi\IEEEauthorrefmark{4}}
\IEEEauthorblockA{Department of Computer Science,
University of Colorado Colorado Springs\\
Email: \IEEEauthorrefmark{1}asolis7@uccs.edu,
\IEEEauthorrefmark{2}jzarkov2@uccs.edu,
\IEEEauthorrefmark{3}jly@uccs.edu,
\IEEEauthorrefmark{4}aatyabi@uccs.edu}}

\maketitle

\begin{abstract}
    The Japanese writing system is complex, with three character types of Hiragana, Katakana, and Kanji. Kanji consists of thousands of unique characters, further adding to the complexity of character identification and literature understanding. Being able to translate handwritten Japanese characters into digital text is useful for data analysis, translation, learning and cultural preservation. In this study, a machine learning approach to analyzing and recognizing handwritten Japanese characters (Kanji) is proposed. The study used an ensemble of three convolutional neural networks (CNNs) for recognizing handwritten Kanji characters and utilized four datasets of MNIST, K-MNIST, Kuzushiji-49 (K49) and the top 150 represented classes in the Kuzushiji-Kanji (K-Kanji) dataset for its performance evaluation. The results indicate feasibility of using proposed CNN-ensemble architecture for recognizing handwritten characters, achieving 99.4\%, 96.4\%, 95.0\% and 96.4\% classification accuracy on MNIST, K-MNIS, K49, and K-Kanji datasets respectively.   
\end{abstract}


\section{Introduction}

    The Japanese writing system is unique compared to the English writing system as it has three different character types of \emph{Hiragana}, \emph{Katakana}, and \emph{Kanji}, each serving a different purpose. In Japanese typing system, Hiragana character type is primarily employed for native Japanese function words while Katakana is commonly utilized for words originating in other languages, and Kanji are used to represent characters with Chinese origin that been adapted to the Japanese language. Hiragana and Katakana both contain 46 basic characters, 71 characters including their diacritics.
    Compared to Kanji, Hiragana and Katakana are relatively simple as they both have characters that point to the same English characters for the sake of translation and pronunciation. 

    Kanji consists of Chinese "Han" characters. These characters first came about from the pictograms of the words they represent. Over time, Kanji became more simplistic until they became the characters they are today.
    While there are many several thousands of Kanji characters, roughly 2000 of them are commonly used.

    \emph{Kuzushiji} is the cursive form of Kanji, and despite being used for over 1000 years, most people fluent in Japanese and/or native to Japan, cannot read Kuzushiji documents anymore. This is partly due to the modernization efforts of the Meiji restoration in 1868, the era were the Japanese education system was reformed, which removed  Kuzushiji from school curriculum. This has left a substantial Japanese literature that can only be worked on depending on availability of a language expert. An estimated three million Japanese pre-modern books and over one billion records and documents use Kuzushiji \cite{ueki2021survey} and this signifies the importance of developing modern approaches to read, analyze, and translate Kuzushiji documents. Aiming to address the need for tools to read and translate this large corpus of knowledge and information that is tied to history of Japanese culture, research community identified Machine Learning (ML) and Artificial Intelligence (AI) as one of the potential mechanism to develop such tools.

\subsection{Problem statement}
As Kuzushiji continues to be phased out by the school curriculum, scientists and researchers must develop reliable systems to aid in the research and cultural preservation of this literature. Given the complex nature of reading Japanese hand-writing, even when AI and ML are used, it is necessary to study mechanism for automated reading these handwritten Japanese characters using AI and ML. In the absence of strong benchmarks on Japanese handwritten character recognition (JHCR), it is important to develop such benchmarks on existing publicly available datasets, specially for Kuzushiji-Kanji.
    
The Kuzushiji-Kanji dataset is one if not the most developed dataset aimed at assisting the development of these AI-empowered automated Japanese hand-writing recognition systems. In this project, we provide an initial benchmark performance measure to be used as a stepping stone for future algorithms and methodology aimed at taking on this dataset while assessing the performance of the proposed CNN-ensemble on other publicly available hand-writing datasets as well. 
    

\subsection{Contribution, Novelty and Outline}
The contribution of this work is the development of Ensemble Classifiers composed of 3 Convolutional Neural Networks (CNNs). 
The feasibility of the proposed CNN-Ensemble method is assessed using four publicly available datasets (see Section \ref{section:Datasets} for more information).

The outline of this study is as follows, a brief literature review of the most recent Japanese and Chinese character recognition methods is presented in section \ref{section:Related_Work}, and description of the datasets used for CNN-Ensemble evaluation are detailed in section \ref{section:Datasets}. Section \ref{section:Methodology} introduces the research methodology followed by results on evaluations on four publicly available datasets in section \ref{section:Results}. Discussion of the results and study conclusion are discussed in sections \ref{section:Results} and \ref{section:Conclusion} respectively.

\section{Related Work}
\label{section:Related_Work}
Japanese Handwritten character recognition (JHCR) is an active research area with challenges associated to complexity and variability of Japanese characters. Multiple mechanisms, including Template-based recognition, convolutional neural networks (CNNs) and recurrent neural networks (RNNs) are considered by the research community to address the task of accurately recognizing handwritten Japanese characters.

Template-based recognition, an approach that is traditionally used for JHCR task, is mainly focused on identifying a best match to a given character from a collection of reference templates gathered in a database. Inability to adequately address and deal-with differences in brush stroke orders and shapes are the main limitations of this approach which convinced researchers to explore and consider the use of more advance computer-vision methods based on machine learning and deep learning.
Convolutional neural networks (CNNs) have become a popular method for JHCR task owing to its ability to automatically learn and extract deep features from images and being able to address limitations of template-based recognition approaches such as stroke order and shape variations. Das et al. \cite{das2014algorithm} reported success in the character recognition of Japan's Hiragana character set, which is similar to Kanji. Rather than examining characters as a whole, in their approach, each character is passed through a pre-processing and normalization stage and is encircled and split into quadrants. in this method, Center of Gravity (CoG) is found for the whole character and for each quadrant. The proposed algorithm produced feature vectors by searching each quadrant for conjunction points (points where multiple strokes intersect) and endpoints and Euclidean distance is measured between the CoG and the significant features of each quadrant. The proposed method achieved 94.1\% recognition rates.
Using Japanese cursive kuzushiji character
image dataset (Kuzushiji-MNIST or K-MNIST) \cite{clanuwat2018deep} and CNN, Ueki and Kojima achieved 73.1\% classification accuracy on JHCR task \cite{icpram20}.

Yang et al. \cite{yang2016dropsample} used the CASIA-OLHWDB 1.0 (DB 1.0) and CASIA-OLHWDB 1.1 (DB 1.1) datasets \cite{liu2011casia} curated by The Institute of Automation, Chinese Academy of Sciences for their Handwritten Chinese Character Recognition (HCCR) study. DB 1.0 dataset is composed of 3740 Chinese characters; each class has 420 examples, and a train-test-split of 336 to 84 is used. The DB 1.1 dataset is composed of 3755 characters; each class has 300 examples, and a train-test-split of 240 to 60 was used. Authors proposed the using combination of a training sampling algorithm based on Leitner's learning box and the integration of a domain-specific knowledge layer in order to improve Chinese Character recognition. Leitner's learning box principle is based on the \emph{"principle of spaced repetition for learning"} which simply states that things that are harder to learn should appear more frequently. Authors used CNN for their CCR task where training samples are assigned a value that helps to place them in one of three bins of well-recognized (these can be sparsely sampled),   confusing (should be sampled more often), and heavily noisy or mislabeled samples (these should rarely be if ever sampled). The proposed approach out-performed state-of-the- art in \cite{liu2013online}, \cite{ciregan2012multi}, and \cite{graham2013sparse} by 2.5\% on the DB 1.0 dataset, achieving classification accuracy of 97.33\% and by 1.4\% on the DB 1.1 dataset, achieving 97.06\% classification accuracy. 


Shi et al. \cite{shi2005radical} used the HITPU handwritten Chinese character database collected by the Harbin Institute of Technology and Hong Kong Polytechnic University in their HCCR study. The dataset comprises 3755 characters from 200 different writers and a total of 751,000 images. Authors used active shape modeling with landmark labeling/recognition and GA for radical recognition. The radical is a fundamental part of Chinese and Japanese character, which significantly aids in recognizing the overall character. The proposed solution achieved a correct radical matching rate of 97.4\%, outperforming state-of-the-art in HCCR (\cite{wang2001optical} and \cite{chung2001complex} ) by roughly 5.9\%.

\section{Datasets}
\label{section:Datasets}

\begin{figure}[!ht]
    \center
    \fboxsep = 1pt 
    \fbox{\includegraphics[width=0.5\textwidth]{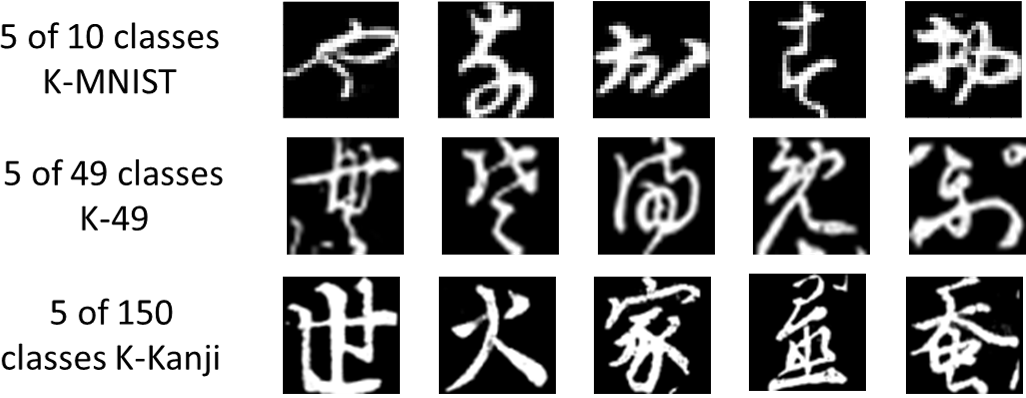}} 
    \caption{Sub-sample of K-MNIST, K-49 and K-Kanji dataset.}
    \label{fig:dataset_example_subset}
\end{figure}

This study uses the datasets curated by \cite{clanuwat2018deep}: \textbf{Kuzushiji-MNIST}, \textbf{Kuzushiji-49} and \textbf{K-Kanji}, referenced as K-MNIST, K-49 and K-Kanji from this point forward, for evaluation and assessment of its proposed CNN-ensemble for JHCR task. 
    
K-MNIST dataset is designed to serve as a replacement for the MNIST dataset, containing 10 classes and following MNIST sample representation format, the K-MNIST samples are represented as $28 \times 28$ grayscale images, total of 70,000 images with a train-test-split of 6 to 1 images per class.
    
K-49 dataset is a much larger datset in comparison to K-MNIST although it is imbalanced. K-49  dataset contains 48 Hiragana characters and one Hiragana iteration mark, total of 49 classes with samples presented as $28 \times 28$ grayscale images with total of 270,920 sample images. The train-test-split in K-49 dataset is 86 to 14 and the distribution is set in such a way to reflect the class imbalance from actual Japanese literature.

The K-Kanji dataset is an imbalanced dataset with total of 140,426 samples represented as $64 \times 64$ grayscale images with a total of 3,832 classes. The sample distribution ranges from 1766 examples in one class to only a single example in another with some classes having up to 3 different representations due to the original Kanji they are derived from which further increases the complexity of this dataset.
In this study, only the top 150 most populated classes in the K-Kanji dataset are used.

Figure \ref{fig:dataset_example_subset} illustrate samples from 5 different classes of each of these three datasets and Table \ref{tab:Kuzushiji_dataset_metadata} presents details about the datsets and their sample sizes and sample distributions in training and testing sets.

\begin{table}
    \centering
    \caption{Metadata of Kuzushiji Datasets}
    \begin{tabular}{|rrrrr|}\hline
    \multicolumn{1}{|c|}{\textbf{\begin{tabular}[c]{@{}c@{}}Dataset\\ Name\end{tabular}}} &
      \multicolumn{1}{|c|}{\textbf{N-Classes}} &
      \multicolumn{1}{|c|}{\textbf{Size}} &
      \multicolumn{1}{|c|}{\textbf{\begin{tabular}[c]{@{}c@{}}Train/Test\\ Split\end{tabular}}} &
      \multicolumn{1}{|c|}{\textbf{\begin{tabular}[c]{@{}c@{}}Class\\ Imbalance\\ (Max-Min)\end{tabular}}} \\\hline
    \multicolumn{1}{|r|}{K-MNIST} & \multicolumn{1}{r|}{10}   & \multicolumn{1}{r|}{70,000}  & \multicolumn{1}{|r|}{6:1} & N/A\\
    \multicolumn{1}{|r|}{K-49}    & \multicolumn{1}{r|}{49}   & \multicolumn{1}{r|}{266,407} & \multicolumn{1}{r|}{7:1} & 6980-392\\
    \multicolumn{1}{|r|}{K-Kanji} & \multicolumn{1}{r|}{3832} & \multicolumn{1}{r|}{140,426} & \multicolumn{1}{r|}{7:3} & 1,766-1 \\\hline
    \end{tabular}
    \label{tab:Kuzushiji_dataset_metadata}
\end{table}

\section{Methodology}
\label{section:Methodology}

\begin{figure*}[!htp]
    \center
    \includegraphics[width=1\linewidth]{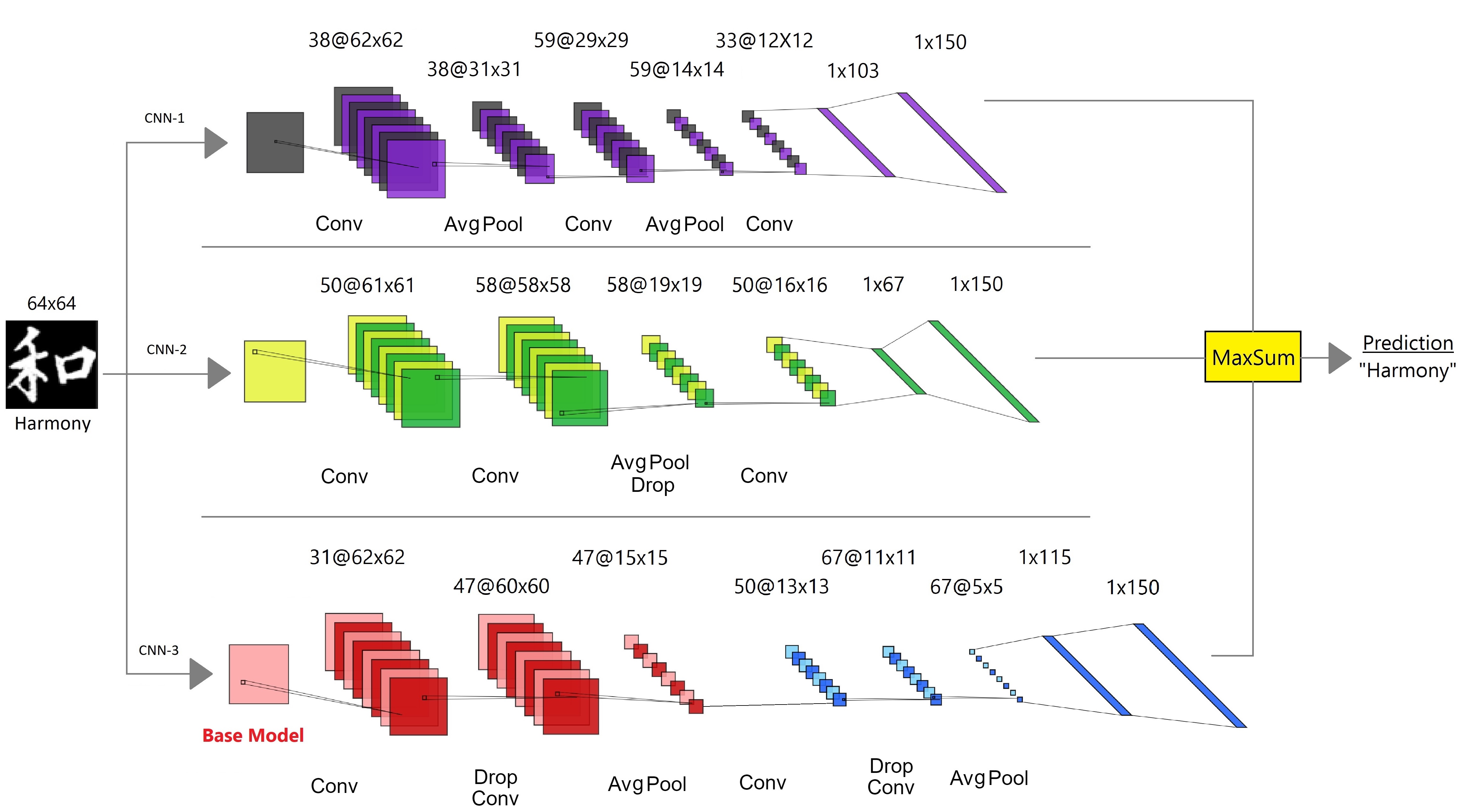}
       \caption{Example CNN ensemble architecture}
    \label{fig:CNN_Ensemble_Model}
\end{figure*}

The ensemble model developed in this study consists of three CNN's as follows:
\begin{enumerate}
    \item CNN-1 is consisted of a convolutional layer, followed by an average pooling layer, a convolutional layer, an average pooling layer, a convolutional layer, a flatten layer, and a densely connected layer.\newline This model is considered in order to attain a more general understanding of input sample shapes. The reduction in sample dimension size from the two average pooling layers makes each sample more generalizable. However, this would be ineffective without the three convolutional layers which help create enough of a distinction between each sample for the model to be effective on its own.
    \item CNN-2 is consisted of two convolutional layers, followed by an average pooling layer, convolutional layer, flatten layer, and a densely connected layer. The architecture is slightly altered and modified for the Kanji dataset by adding a convolutional layer before the flatten layer.\newline This model uses the three convolutional layers to focus its learning on the specific shape of each sample while the pooling layer is used to reduce computational complexity.
    \item CNN-3 is the transfer learning model. This model's architecture utilizes a base model that is pre-trained on another dataset. The base model is then used as the head of the transfer learning model before additional layers are appended to it. The mechanism used for transfer learning in CNN-3 model depend on the dataset being used for evaluation (e.g., KMNIST, K49, or Kanji dataset). That is, when CNN-3 model is to be used for evaluating samples in KMNIST dataset, the base model is pre-trained using MNIST dataset and when K-49 is to be used for evaluating the model performance, the CNN-3 base model is pre-trained using KMNIST dataset. Finally, CNN-3 base model is pre-trained by samples of K-49 dataset when the evaluation is done by samples of Kanji dataset.
    \begin{enumerate}
        \item Base Model architecture includes a convolutional layer, followed by a dropout layer, convolutional layer, flatten layer, and a densely connected layer. The architecture is modified for the Kanji dataset by adding an average pooling layer before the flatten layer. This modification is considered due to samples of the Kanji dataset being much larger in size compared to the samples of the other datasets (64$\times$64 pixels in K-Kanji dataset versus 28$\times$28 pixels in K-MNIST and K-49). The extra average pooling layer is placed to accommodate for the image size differences, reducing computational cost while increasing the accuracy.
        \item Additional layers: In CNN-3, the layers of the base model are followed by additional layers of convolutional layer, dropout layer, convolutional layer, average pooling layer, flatten layer, and a densely connected layer. Pre-training the base model allow the model to recognize general image fundamentals. Freezing the weights in the base model helps to retain the knowledge of image fundamentals while the new layers are tuned on the new target domain (training set samples from evaluating dataset). Starting with basic knowledge generally allows transfer learning models to be faster and more accurate than regular models.
        \end{enumerate}
\end{enumerate}
Each CNN model (CNN1, CNN-2, and CNN-3) in CNN-Ensemble architecture is connected to a final output layer to allow the accuracy of the output to be evaluated. An illustration of the CNN-ensemble architecture is presented in figure \ref{fig:CNN_Ensemble_Model}.

In this study, the proposed ensemble model is a conglomeration (rather than combination) of CNN-models. As expected from ensemble models, single CNN model architectures are expected to be less accurate in their predictions and conglomerating the models can mitigate single model weaknesses through group voting. The proposed ensemble architecture utilizes the strengths of three different CNN-models to offset incorrect individual predictions and increase overall validation accuracy on each dataset. 

The performance of the proposed CCN-ensemble architecture is assessed on all three datasets used in this study (MNIST, K-MNIST, K-49 and K-Kanji), one dataset at the time. Each CNN-model in the proposed CNN-ensemble is trained  and validated individually (using the same training and validation sets on each dataset for all three CNN-models) and a voting mechanism based on aggregating votes of all three CNN-models into a single vector and taking the $maxsum$ of the final vector is used to generate the overall ensemble prediction on the validation set of each dataset.

The results of the individual models and the CNN-ensemble model are presented in table \ref{tab:individual_vs_ensemble_table} and further elaborated in section \ref{section:Results}.

\section{Results}
\label{section:Results}

\subsection{MNIST Results}
The three CNN models were tested on the MNIST dataset to gain a performance baseline. 
\begin{itemize}
    \item CNN-1: been able to achieve a 99.00\% accuracy on the validation data.
    \item CNN-2 been able to achieve a 97.30\% accuracy on the validation data.
    \item CNN-3 been able to achieve a 99.03\% accuracy on the validation data. The MNIST dataset was used as a benchmark for the other datasets so it did not use any transfer learning.
    \item CNN-Ensemble produce a 99.35\% accuracy on the validation data. The  average classification performance across CNN models achieved classification accuracy of 98.35\%, indicating that using an ensemble model improved the average model accuracy by 1.0\%.
\end{itemize}
       
These results can also be found on table \ref{tab:individual_vs_ensemble_table}.
\subsection{K-MNIST Results}
The three CNN models been evaluated on the K-MNIST dataset aiming to provide  performance comparison between the ensemble model and individual CNN models. The results are as follows:
\begin{itemize}
    \item CNN-1 achieved a 94.44\% accuracy on the validation data.
    \item CNN-2 achieved a 95.11\% accuracy on the validation data.
    \item CNN-3 achieved a 95.15\% accuracy on the validation data. Samples of  MNIST dataset been used to pre-train the base model in CNN-3 although marginal performance improvement is observed compare to CNN-1 and CNN-2.
    \item CNN-Ensemble achieved classification accuracy of 96.37\% on the validation data, showing an improvement of 1.47\% on average performance achieved across the three CNN models (94.90\%) while also outperforming individual CNN models' performances.
\end{itemize}
These results are also illustrated on table \ref{tab:individual_vs_ensemble_table}.

\subsection{K-49 Results}
All three CNN models are also tested on the K-49 dataset, with model 3 using transfer learning from the K-MNIST dataset. The results are as follows:
\begin{itemize}
    \item CNN-1 achieved a 91.80\% accuracy on the validation data.
    \item CNN-2 achieved a 93.01\% accuracy on the validation data.
    \item CNN-3 achieved a 93.09\% accuracy on the validation data. CNN-3 only marginally outperformed CNN models 1 and 2 on the K-49 dataset using transfer learning from the K-MNIST dataset. However, the average training time for the transfer learning model was reduced from 1900 seconds to 987 seconds, or by 48\%.
    \item CNN-Ensemble achieved classification accuracy of 95.04\% on the validation data. This model outperformed both average CNN models performances (92.63\%, showing 2.41\% performance improvement) and individual CNN models performances.
\end{itemize}
These results can additionally be found on table \ref{tab:individual_vs_ensemble_table}.
    
\subsection{K-Kanji Results}
K-Kanji dataset is considered as the true target class of this study since, to the best of our knowledge, its performance been never evaluated in literature. 
This dataset consists of a total of 3832 Kanji characters in a 64$\times$64 grayscale images of Kanji handwriting. This dataset consists of a smaller set of 140,426 images and is highly imbalanced where numbers of examples per class can range from 1,766 to even just a single example. For evaluation of the dataset and to assess feasibility of the proposed CNN-Ensemble architecture, only the top 150 populated character classes are considered in this study.
\begin{itemize}
    \item CNN-1 achieved 92.83\% classification accuracy on the validation set.
    \item CNN-2 achieved 93.49\% classification accuracy on the validation set.
    \item CNN-3 achieved 95.01\% classification accuracy on the validation set. The base model in CNN-3 is pre-trained using samples from K-49 dataset. The results indicate the substantial improvement in performance on K-Kanji dataset compared to CNN-1 and CNN-2 models. This is likely due to the increased sample size in K-49 dataset (266,407 samples, see table \ref{tab:Kuzushiji_dataset_metadata} for more details) that better accommodated learning through transfer learning (base model in CNN-3), allowing the model to learn more information about the images in the base model.
    \item CNN-Ensemble achieved 96.43\% classification accuracy on the validation data. The average classification performance across 3 CNN models achieved 93.77\% classification accuracy. This indicates improvement of 2.65\% in CNN-Ensemble compared to average model and outperforming all 3 CNN models. This is likely due to an increased sample size compared to the KMNIST and K-49 datasets. The base model for Kanji included an additional pooling layer because of the increased sample size. This extra layer is likely the contributing factor to the great success of the Kanji transfer learning model.
\end{itemize}
A summary of these results are presented in table \ref{tab:individual_vs_ensemble_table}.

\begin{table}[!ht]
    \centering
    \begin{scriptsize}
    \caption{Individual CNN model performance compared against Ensemble Model.}
    \begin{tabular}{lccccc}
    \multicolumn{1}{c}{} &CNN 1 &CNN 2 &CNN 3 &
      \begin{tabular}[c]{@{}c@{}}Model\\ Average\end{tabular} &
      \begin{tabular}[c]{@{}c@{}}Ensemble\\ Model\end{tabular} \\ \hline
    \multicolumn{1}{|l|}{MNIST} &99.00\% &97.30\% &99.03\% &98.35\% &
      \multicolumn{1}{c|}{\textbf{99.35\%}} \\
    \multicolumn{1}{|l|}{K-MNIST} &94.44\% &95.11\% &95.15\% &94.90\% & 
     \multicolumn{1}{c|}{\textbf{96.37\%}} \\
    \multicolumn{1}{|l|}{K-49}&91.80\% &93.01\% &93.09\% &92.63\% &  
     \multicolumn{1}{c|}{\textbf{95.04\%}} \\
    \multicolumn{1}{|l|}{K-Kanji} & 92.83\% & 93.49\% & 95.01\% & 93.77\% & 
      \multicolumn{1}{c|}{\textbf{96.43}\%}  \\ \hline
    \end{tabular}
    \label{tab:individual_vs_ensemble_table}
    \end{scriptsize}
\end{table}

\begin{table}[!ht]
    \caption{Comparison of CNN-Ensemble performance with State-of-the-art across datasets utilized in the study.}
    \begin{scriptsize}
    \begin{tabular}{|p{3.5cm}|p{.35cm}p{1.05cm}p{.6cm}p{.9cm}}
     \multicolumn{1}{c}{}&MNIST &K-MNIST &K-49&K-Kanji\\\hline
    4-Nearest Neighbor Baseline \cite{clanuwat2018deep}&
      \multicolumn{1}{c}{97.14\%} &\multicolumn{1}{c}{91.56\%} &
      \multicolumn{1}{c}{86.01\%} &\multicolumn{1}{c|}{-} \\
    Keras Simple CNN Benchmark \cite{clanuwat2018deep}&
      \multicolumn{1}{c}{99.06\%} &\multicolumn{1}{c}{95.12\%} &
      \multicolumn{1}{c}{89.25\%} &\multicolumn{1}{c|}{-} \\
    PreActResNet-18 \cite{he2016identity}&
      \multicolumn{1}{c}{99.56\%} &\multicolumn{1}{c}{97.82\%} &
      \multicolumn{1}{c}{96.64\%} &\multicolumn{1}{c|}{-} \\
    PreActResNet-18 + Input Mixup \cite{zhang2017mixup}&
      \multicolumn{1}{c}{99.54\%} &\multicolumn{1}{c}{98.41\%} &
      \multicolumn{1}{c}{97.04\%} &\multicolumn{1}{c|}{-} \\
    PreActResNet-18 + Manifold Mixup \cite{verma2019manifold}&
      \multicolumn{1}{c}{99.54\%} &\multicolumn{1}{c}{98.41\%} &
      \multicolumn{1}{c}{97.04\%} &\multicolumn{1}{c|}{-} \\
   \textbf{CNN-Ensemble}&
      \multicolumn{1}{c}{99.35\%} &\multicolumn{1}{c}{96.37\%} &
      \multicolumn{1}{c}{95.04\%} &\multicolumn{1}{c|}{96.43\%} \\ \hline
    \end{tabular}
   \end{scriptsize}
   \label{tab:model_results_validation_table}
\end{table}

\section{Discussion}
\label{section:Discussion}
In table \ref{tab:model_results_validation_table}, no state-of-the-art performance is provided for K-Kanji dataset since, to the best of our knowledge, this dataset never been considered by research community, possibly due to complexities associated with the dataset, specially the imbalance nature of the dataset which makes it more difficult to effectively train models on smaller sampled character classes.

Table \ref{tab:model_results_validation_table} is augmented from the \cite{dutta2018improving} study to show how a variety of algorithms perform on the MNIST, K-MNIST and K-49 dataset.
This table has been updated to show the result of the ensemble architecture and it's performance on the same datasets with the addition of the K-Kanji dataset.


To better understand the performance variations across the 150 classes of the K-Kanji dataset, the testing dataset F1-Scores for each class is illustrated in table \ref{tab:F1_Score_Table}. The F1-Scores are presented in descending order, ranging from F1-Score of 1 to 0.49. It is noticeable that majority of the classes are lacking adequate number of samples for training and tuning of CNN models which speaks to difficulties in using K-Kanji dataset.

\begin{table*}[!ht]
  \centering
  \caption{Multi-class classification F1 Scores for top 150 K-Kanji Characters. "\# Val. Samp." and "\# Tran. Samp." represent number of validation and training samples respectively.}
  \begin{scriptsize}
    \setlength{\tabcolsep}{4.0pt}
    \begin{tabular}{llll|llll|llll|llll}
      \multicolumn{1}{c}{Kanji} &
        \multicolumn{1}{c}{\begin{tabular}[c]{@{}c@{}}F1.\\ Score\end{tabular}} &
        \multicolumn{1}{c}{\begin{tabular}[c]{@{}c@{}}\# Val. \\ Samp.\end{tabular}} &
        \multicolumn{1}{c|}{\begin{tabular}[c]{@{}c@{}}\# Tran. \\ Samp.\end{tabular}} &
        \multicolumn{1}{c}{Kanji} &
        \multicolumn{1}{c}{\begin{tabular}[c]{@{}c@{}}F1.\\ Score\end{tabular}} &
        \multicolumn{1}{c}{\begin{tabular}[c]{@{}c@{}}\# Val. \\ Samp.\end{tabular}} &
        \multicolumn{1}{c|}{\begin{tabular}[c]{@{}c@{}}\# Tran. \\ Samp.\end{tabular}} &
        \multicolumn{1}{c}{Kanji} &
        \multicolumn{1}{c}{\begin{tabular}[c]{@{}c@{}}F1.\\ Score\end{tabular}} &
        \multicolumn{1}{c}{\begin{tabular}[c]{@{}c@{}}\# Val. \\ Samp.\end{tabular}} &
        \multicolumn{1}{c|}{\begin{tabular}[c]{@{}c@{}}\# Tran. \\ Samp.\end{tabular}} &
        \multicolumn{1}{c}{Kanji} &
        \multicolumn{1}{c}{\begin{tabular}[c]{@{}c@{}}F1.\\ Score\end{tabular}} &
        \multicolumn{1}{c}{\begin{tabular}[c]{@{}c@{}}\# Val. \\ Samp.\end{tabular}} &
        \multicolumn{1}{c}{\begin{tabular}[c]{@{}c@{}}\# Tran. \\ Samp.\end{tabular}} \\ \hline
      male        & 1.0   & 71  & 170  & west                 & 0.960 & 101 & 232  & six             & 0.937 & 134 & 299 & honorable    & 0.912 & 180 & 372 \\
      one         & 0.992 & 550 & 1218 & creek                & 0.960 & 98  & 221  & day             & 0.937 & 294 & 703 & behind       & 0.912    & 126 & 295 \\
      egg         & 0.990 & 55  & 155  & below                & 0.960 & 140 & 304  & hand            & 0.937   & 161 & 366 & long       & 0.912    & 66  & 159 \\
      or\_again   & 0.987 & 400 & 1051 & husband              & 0.960 & 100 & 231  & inside          & 0.936 & 109 & 267 & tree         & 0.909 & 120 & 267 \\
      rice\_field & 0.984 & 66  & 159  & house                & 0.959 & 121 & 269  & same            & 0.936 & 151 & 358 & say\_word    & 0.906  & 151 & 353 \\
      silkworm    & 0.983 & 151 & 369  & matter               & 0.959 & 515 & 1159 & extent          & 0.936 & 61  & 159 & sake         & 0.905 & 146 & 335 \\
      capital     & 0.982 & 58  & 160  & life                 & 0.958 & 121 & 268  & country\_modern & 0.936 & 197 & 471 & eye          & 0.898 & 106 & 266 \\
      see         & 0.982 & 319 & 714  & somebody             & 0.958 & 186 & 406  & think           & 0.935 & 118 & 266 & color        & 0.894 & 124 & 287 \\
      spirit      & 0.982 & 55  & 153  & good\_luck           & 0.958 & 151 & 369  & east            & 0.935 & 119 & 265 & old          & 0.890 & 62  & 159 \\
      soup        & 0.982 & 140 & 324  & seven                & 0.958 & 83  & 180  & possess         & 0.935 & 193 & 434 & climate      & 0.890 & 217 & 511 \\
      say         & 0.981 & 523 & 1243 & reside               & 0.957 & 116 & 265  & to\_be\_        & 0.935 & 290 & 630 & springtime   & 0.881 & 105 & 247 \\
      pull        & 0.976 & 104 & 241  & mountain             & 0.955 & 163 & 366  & gold            & 0.933 & 73  & 170 & private      & 0.879  & 55  & 150 \\
      three       & 0.974 & 238 & 551  & enter                & 0.954 & 381 & 878  & son             & 0.933 & 52  & 153 & grass        & 0.879 & 40  & 158 \\
      above       & 0.974  & 255 & 610  & time                 & 0.953 & 181 & 397  & wind            & 0.932 & 124 & 296 & flower      & 0.878 & 93  & 219 \\
      little      & 0.973 & 270 & 628  & place                & 0.953 & 187 & 427  & month           & 0.930 & 149 & 350 & placement    & 0.878 & 100 & 229 \\
      just\_so    & 0.973 & 248 & 545  & reach\_out           & 0.953  & 73  & 172  & word            & 0.930 & 48  & 152 & salt        & 0.876 & 77  & 179 \\
      gods        & 0.972 & 72  & 170  & now                  & 0.952 & 182 & 401  & in\_front       & 0.927 & 196 & 471 & traffic      & 0.875 & 78  & 179 \\
      adhere      & 0.972 & 90  & 186  & eat                  & 0.951  & 82  & 181  & four            & 0.926  & 124 & 292 & of         & 0.874 & 148 & 343 \\
      thing       & 0.972 & 360 & 839  & mulberry             & 0.951  & 79  & 184  & few             & 0.925  & 60  & 158 & take       & 0.873  & 123 & 280 \\
      what        & 0.971  & 251 & 569  & have\_the\_honor\_to & 0.949 & 139 & 298  & mouth           & 0.925 & 147 & 339 & water       & 0.872 & 209 & 512 \\
      woman       & 0.971 & 208 & 470  & generation           & 0.949 & 171 & 379  & paper           & 0.924 & 95  & 220 & utilize      & 0.865 & 68  & 159 \\
      some        & 0.971 & 51  & 153  & in                   & 0.947 & 216 & 507  & exit            & 0.924 & 325 & 727 & section      & 0.864 & 56  & 155 \\
      door        & 0.971 & 122 & 273  & ego                  & 0.947 & 70  & 165  & name            & 0.924 & 127 & 300 & negative     & 0.862 & 76  & 177 \\
      person      & 0.970 & 464 & 1111 & child                & 0.946 & 313 & 706  & someone         & 0.924 & 68  & 162 & plains       & 0.862 & 73  & 169 \\
      ten         & 0.968 & 140 & 319  & Esq                  & 0.946 & 160 & 364  & fish            & 0.922 & 77  & 180 & part         & 0.853 & 111 & 265 \\
      left        & 0.968 & 80  & 183  & roof                 & 0.945 & 92  & 216  & leaf            & 0.922 & 74  & 178 & plump        & 0.852 & 51  & 152 \\
      this        & 0.965 & 340 & 781  & eight                & 0.945 & 78  & 182  & outside         & 0.920 & 79  & 183 & year         & 0.847 & 84  & 184 \\
      interval    & 0.965 & 88  & 186  & state                & 0.945 & 98  & 218  & tea             & 0.919  & 89  & 187 & jewel       & 0.843 & 52  & 152 \\
      harmony     & 0.965 & 86  & 186  & bake                 & 0.944 & 70  & 163  & turn\_into      & 0.918 & 74  & 174 & soil         & 0.842 & 58  & 155 \\
      two         & 0.964 & 192 & 433  & white                & 0.944 & 98  & 217  & heart           & 0.918 & 195 & 468 & country\_old & 0.841 & 77  & 176 \\
      heavens     & 0.964 & 69  & 163  & stand\_up            & 0.944 & 156 & 366  & logic           & 0.917 & 91  & 195 & No           & 0.836 & 47  & 152 \\
      species     & 0.962 & 79  & 184  & fit                  & 0.944    & 193 & 444  & fire            & 0.917 & 55  & 153 & write     & 0.824 & 79  & 182 \\
      direction   & 0.962 & 262 & 608  & salary               & 0.943 & 167 & 370  & make            & 0.915 & 68  & 160 & even         & 0.810 & 57  & 154 \\
      bird        & 0.962 & 105 & 258  & reign                & 0.941 & 99  & 227  & hear            & 0.915 & 111 & 269 & book         & 0.794 & 65  & 157 \\
      that        & 0.962 & 308 & 694  & night                & 0.941 & 121 & 267  & going           & 0.914 & 92  & 212 & come         & 0.651 & 108 & 264 \\
      attend      & 0.961 & 129 & 302  & boil                 & 0.940 & 73  & 171  & five            & 0.913 & 168 & 378 & due          & 0.490 & 33  & 164 \\
      flavor      & 0.961 & 119 & 267  & large                & 0.939  & 332 & 734  & road            & 0.913 & 93  & 221 &             &          &     &     \\
      cut         & 0.961 & 91  & 203  & right                & 0.938 & 105 & 248  & bright          & 0.912 & 57  & 156 &              &          &     &    
    \end{tabular}
  \label{tab:F1_Score_Table}
  \end{scriptsize}
\end{table*}

A \underbar{SH}apely \underbar{A}dditive ex\underline{P}lanations (SHAP) analysis is performed on each CNN model used in CNN-Ensemble to better understand the results achieved. SHAP analysis is a methodology used to represent contributions of different features in a given sample in the prediction achieved by a machine learning model \cite{lundberg2017unified}. SHAP values, ranging from -1 (commonly represented by blue color in SHAP images) to 1 (commonly represented by red color in SHAP images), indicates the impact of each feature on prediction, by higher SHAPE units indicating higher contribution in achieved prediction. SHAP values are calculate based on acquiring the  difference between prediction achieved by machine learning model for a given sample with and without using the feature. 
    
\begin{figure*}[!ht]
  \center
  \fboxsep = 1pt 
  \fbox{\includegraphics[width=1.0\textwidth]{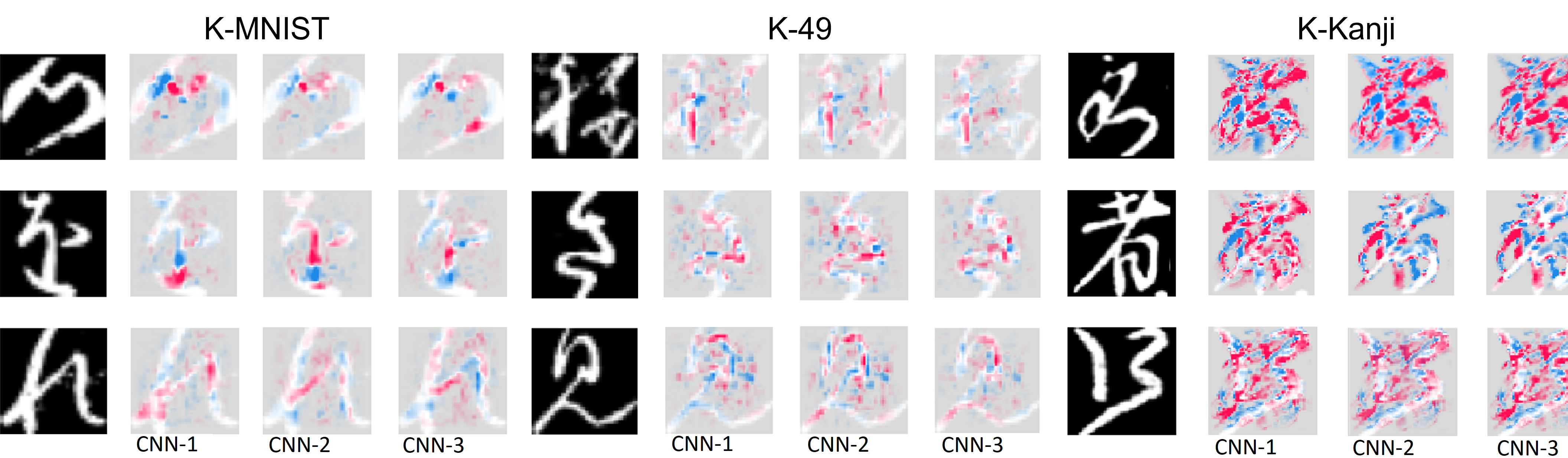}} 
  \caption{SHAP Analysis Output of CNN1, CNN-2 and CNN-3 models on three samples for K-MNIST, K-49 and K-Kanji datasets.}
  \label{fig:shap_analysis_plot}
\end{figure*}

Figure \ref{fig:shap_analysis_plot} illustrates the results of SHAP analysis on the three CNN models used in this study using three samples, randomly chosen, from each dataset considered in the study, e.g., KMNIST, k-49, and K-Kanji. Brighter red pixels indicate a positive contribution to the classification while the blue pixels indicate negative contribution to the model prediction. The results highlights little to no discernible difference between the models which is an indication of the observed inability of CNN-ensemble model in achieving substantial gain in classification performance compared to individual CNN models, e.g., CNN-1, CNN-2, and CNN-3.
    

\section{Conclusion}
\label{section:Conclusion}

This study discussed the development of a machine learning approach to identify and translate handwritten Japanese characters (Kanji) into its corresponding English translation. It is imperative to develop reliable systems for aiding in the research and cultural preservation of Kuzushiji literature given that very few individuals are able to translate these documents with millions of such documents still in process of being translated and prepared for use by researches interested in understanding historical context withhold within them.

This study employed an ensemble of three convolutional neural networks (CNNs) and evaluated its performance. One of the aims and main contributions of this study is to develop benchmarks for Japanese handwritten character recognition, using publicly available datasets. To this end, four publicly available datasets of Kuzushiji-
MNIST, Kuzushiji-49 and K-Kanji, referred to as as K-MNIST,
K-49 and K-Kanji, are used to evaluate the performance of the proposed CNN-Ensemble architecture. The results indicated feasibility of the proposed CNN-Ensemble architecture in correctly predicting various Japanese Handwritten characters, achieving average accuracy of 96\% across the datasets used, with individual CNN models showing a range of performance from 91.8\% (CNN-1 on K-49 dataset) to 95.15\% (CNN-3 on K-MNIST dataset). 

    

\bibliographystyle{IEEEtran} \bibliography{references} 

\begin{thebibliography}{10}
\providecommand{\url}[1]{#1}
\csname url@samestyle\endcsname
\providecommand{\newblock}{\relax}
\providecommand{\bibinfo}[2]{#2}
\providecommand{\BIBentrySTDinterwordspacing}{\spaceskip=0pt\relax}
\providecommand{\BIBentryALTinterwordstretchfactor}{4}
\providecommand{\BIBentryALTinterwordspacing}{\spaceskip=\fontdimen2\font plus
\BIBentryALTinterwordstretchfactor\fontdimen3\font minus
  \fontdimen4\font\relax}
\providecommand{\BIBforeignlanguage}[2]{{%
\expandafter\ifx\csname l@#1\endcsname\relax
\typeout{** WARNING: IEEEtran.bst: No hyphenation pattern has been}%
\typeout{** loaded for the language `#1'. Using the pattern for}%
\typeout{** the default language instead.}%
\else
\language=\csname l@#1\endcsname
\fi
#2}}
\providecommand{\BIBdecl}{\relax}
\BIBdecl

\bibitem{ueki2021survey}
K.~Ueki and T.~Kojima, ``Survey on deep learning-based kuzushiji recognition,''
  in \emph{International Conference on Pattern Recognition}.\hskip 1em plus
  0.5em minus 0.4em\relax Springer, 2021, pp. 97--111.

\bibitem{das2014algorithm}
S.~Das and S.~Banerjee, ``An algorithm for japanese character recognition,''
  \emph{International Journal of Image, Graphics and Signal Processing},
  vol.~7, no.~1, p.~9, 2014.

\bibitem{clanuwat2018deep}
T.~Clanuwat, M.~Bober-Irizar, A.~Kitamoto, A.~Lamb, K.~Yamamoto, and D.~Ha.
  (2018) Deep learning for classical japanese literature.

\bibitem{icpram20}
K.~Ueki and T.~Kojima, ``Japanese cursive character recognition for efficient
  transcription,'' in \emph{Proceedings of the 9th International Conference on
  Pattern Recognition Applications and Methods - Volume 1: ICPRAM,},
  INSTICC.\hskip 1em plus 0.5em minus 0.4em\relax SciTePress, 2020, pp.
  402--406.

\bibitem{yang2016dropsample}
W.~Yang, L.~Jin, D.~Tao, Z.~Xie, and Z.~Feng, ``Dropsample: A new training
  method to enhance deep convolutional neural networks for large-scale
  unconstrained handwritten chinese character recognition,'' \emph{Pattern
  Recognition}, vol.~58, pp. 190--203, 2016.

\bibitem{liu2011casia}
C.-L. Liu, F.~Yin, D.-H. Wang, and Q.-F. Wang, ``Casia online and offline
  chinese handwriting databases,'' in \emph{2011 international conference on
  document analysis and recognition}.\hskip 1em plus 0.5em minus 0.4em\relax
  IEEE, 2011, pp. 37--41.

\bibitem{liu2013online}
\BIBentryALTinterwordspacing
------, ``Online and offline handwritten chinese character recognition:
  Benchmarking on new databases,'' \emph{Pattern Recognition}, vol.~46, no.~1,
  pp. 155--162, 2013. [Online]. Available:
  \url{https://www.sciencedirect.com/science/article/pii/S0031320312002919}
\BIBentrySTDinterwordspacing

\bibitem{ciregan2012multi}
D.~Ciregan, U.~Meier, and J.~Schmidhuber, ``Multi-column deep neural networks
  for image classification,'' in \emph{2012 IEEE conference on computer vision
  and pattern recognition}.\hskip 1em plus 0.5em minus 0.4em\relax IEEE, 2012,
  pp. 3642--3649.

\bibitem{graham2013sparse}
B.~Graham, ``Sparse arrays of signatures for online character recognition,''
  \emph{arXiv preprint arXiv:1308.0371}, 2013.

\bibitem{shi2005radical}
D.~Shi, G.~Ng, R.~Damper, and S.~Gunn, ``Radical recognition of handwritten
  chinese characters using ga-based kernel active shape modelling,'' \emph{IEE
  Proceedings-Vision, Image and Signal Processing}, vol. 152, no.~5, pp.
  634--638, 2005.

\bibitem{wang2001optical}
A.-B. Wang and K.-C. Fan, ``Optical recognition of handwritten chinese
  characters by hierarchical radical matching method,'' \emph{Pattern
  Recognition}, vol.~34, no.~1, pp. 15--35, 2001.

\bibitem{chung2001complex}
F.-L. Chung and W.~W. Ip, ``Complex character decomposition using deformable
  model,'' \emph{IEEE Transactions on Systems, Man, and Cybernetics, Part C
  (Applications and Reviews)}, vol.~31, no.~1, pp. 126--132, 2001.

\bibitem{he2016identity}
K.~He, X.~Zhang, S.~Ren, and J.~Sun, ``Identity mappings in deep residual
  networks,'' in \emph{Computer Vision--ECCV 2016: 14th European Conference,
  Amsterdam, The Netherlands, October 11--14, 2016, Proceedings, Part IV
  14}.\hskip 1em plus 0.5em minus 0.4em\relax Springer, 2016, pp. 630--645.

\bibitem{zhang2017mixup}
H.~Zhang, M.~Cisse, Y.~N. Dauphin, and D.~Lopez-Paz, ``mixup: Beyond empirical
  risk minimization,'' \emph{arXiv preprint arXiv:1710.09412}, 2017.

\bibitem{verma2019manifold}
V.~Verma, A.~Lamb, C.~Beckham, A.~Najafi, I.~Mitliagkas, D.~Lopez-Paz, and
  Y.~Bengio, ``Manifold mixup: Better representations by interpolating hidden
  states,'' in \emph{International conference on machine learning}.\hskip 1em
  plus 0.5em minus 0.4em\relax PMLR, 2019, pp. 6438--6447.

\bibitem{dutta2018improving}
K.~Dutta, P.~Krishnan, M.~Mathew, and C.~Jawahar, ``Improving cnn-rnn hybrid
  networks for handwriting recognition,'' in \emph{2018 16th international
  conference on frontiers in handwriting recognition (ICFHR)}.\hskip 1em plus
  0.5em minus 0.4em\relax IEEE, 2018, pp. 80--85.

\bibitem{lundberg2017unified}
S.~M. Lundberg and S.-I. Lee, ``A unified approach to interpreting model
  predictions,'' \emph{Advances in neural information processing systems},
  vol.~30, 2017.

\end{thebibliography}

\end{document}